\documentclass[11pt]{article}
\usepackage[preprint]{acl}
\usepackage{times}
\usepackage{latexsym}
\usepackage{graphicx}
\usepackage{booktabs}
\usepackage{amsmath}
\usepackage{xcolor}
\usepackage{xeCJK}
\newcommand{\zh}[1]{#1}
\newcommand{\ours}{L\"ushi-1840}
\newcommand{\old}{L\"ushi-old}

\title{\ours{}: Building and Probing a Knowledge Cutoff\\for a Classical Chinese Language Model}

\author{Alex Liu \\ Ybuild LLC \\ \texttt{alex@ybuild.ai}}

\begin{document}
\maketitle

\begin{abstract}
Language models trained only on text from before a given year are useful for studying hindsight and
historical language, but the open collections of pre-modern Chinese carry no publication dates and
unreliable dynasty labels. We present \ours{}, a 983M-parameter model trained from scratch on 12{,}262
books (1.28B characters) selected as composed before 1840. A first corpus, dated by a common heuristic, in
effect ended at 1644 while containing twentieth-century material. We describe the re-dating and cleaning
pipeline that replaced it and an audit with markers the pipeline did not use, which still finds later
material. On the model side we measure surprisal, relative to a modern reference, on annals
compiled after the cutoff. For events after 1840 against events up to 1820 the ratio is 1.38 (95\%
interval 1.33--1.44), against about 1.00 for modern general-purpose models. In a series of 294M models,
adding the later court records lowers the ratio from 1.33 to 1.16, and the two
models diverge from the 1820s, where the added records begin. The probe tracks the coverage of
sources and does not establish a single cutoff year; none of our probes exposes the audited
contamination. Model, book list and scripts will be released.
\end{abstract}

\section{Introduction}

A language model whose training data ends at a known year is an instrument: it can be asked what was
foreseeable from inside a period, used as a control for look-ahead bias, or studied as a sample of a
period's written language. Several such \emph{vintage} models have been trained from scratch in the last
two years~\citep{luo2026typewriter,levine2026talkie,goettlich2025history,hla2026machina,grigorian2026timecapsule};
the documented ones are, to our knowledge, English, with cutoffs between 1875 and 1946. Their corpora are scanned library
books, which come with a year of publication. The cutoff is obtained by filtering on that year, and the
remaining work goes into removing modern metadata; serials dated by the year the series began are a
reported source of leakage~\citep{luo2026typewriter}.

Pre-modern Chinese is an obvious next case and a harder one. Literary Chinese was the written standard for
some two thousand years and the surviving corpus is large and already transcribed. But the open collections
behind most Classical Chinese models~\citep{garychowcmu2018daizhige,wittern2014kanripo} are organised by
bibliographic category, not by date. Files carry a title and at best a dynasty; many are twentieth-century
critical editions in which modern notes are interleaved with the old text in the same register; and because
the written language changed slowly, a preface of 1890 does not look different from one of 1790. Existing
Classical Chinese models make no claim about a cutoff~\citep{chen2026internal,zhao2019uer,liu2023sikugpt},
and one describes its coverage as extending to the Republican period~\citep{yao2025wenyangpt}.

We set out to train a model on text composed before 1840, the conventional start of modern history in
Chinese historiography (the First Opium War). The first attempt failed in an instructive
way. Dating books by the first dynasty name in their opening lines produced a corpus that was, on
inspection, a pre-1644 corpus (Qing text: about 1.5\%) with late contamination scattered through it; the
resulting model, asked to continue the list of Qing reign names up to the one current in 1840, proposed the
next one (Figure~\ref{fig:samples}). This paper reports what it took to repair this, how far the repair can
be checked, and where the checks stop working. We keep the target year in the model's name;
on the probe of \S\ref{sec:verify} the observed boundary of coverage is nearer 1820.

\begin{figure}[t]
\small
\begin{tabular}{@{}p{0.98\columnwidth}@{}}
\toprule
\textbf{Prompt:} \zh{順治、康熙、雍正、乾隆、嘉慶、道光、}\\
{\footnotesize (the Qing reign names in order, up to Daoguang, 1821--1850)}\\
\old{}: \zh{咸豐、光緒、崇禎、同治、光緒}\ldots\\
{\footnotesize \emph{Xianfeng} (1851--61), \emph{Guangxu} (1875--1908), \ldots, \emph{Tongzhi} (1862--74)}\\
\ours{}: \zh{天啓、崇禎、皇清、雍正、乾隆}\ldots\\
{\footnotesize two late-Ming reign names, then back to the early Qing}\\
\midrule
\textbf{Prompt:} \zh{火輪車者，} {\footnotesize (``The fire-wheel carriage is''; a mid-nineteenth-century word for the train)}\\
\old{}: \zh{即今之火車也} {\footnotesize (``what is now called a train'')}\\
\ours{}: \zh{以其有火力，故名火輪} {\footnotesize (``so named because it has the force of fire'')}\\
\bottomrule
\end{tabular}
\caption{Greedy continuations from the model trained on the heuristically dated corpus (\old{}) and on the
re-dated corpus (\ours{}); two of fourteen prompts, chosen for illustration. Both models were intended to
have an 1840 cutoff.}
\label{fig:samples}
\end{figure}

Our contributions are:

\begin{itemize}
\setlength\itemsep{0pt}
\item \textbf{A model and corpus.} \ours{} (983M parameters, character-level) and the list of 12{,}262
books (1.28B characters) behind it. We are not aware of another model of pre-modern Chinese whose
cutoff has been audited on both the corpus and the model side. One training run takes 21 hours on a single H100.
\item \textbf{A documented failure and repair.} A pipeline for corpora without publication dates,
reported with before/after audits, including an audit with markers the pipeline did not use, which finds
residual late material that we list (\S\ref{sec:corpus}).
\item \textbf{A model-side probe with controls.} Relative surprisal on annals compiled after the cutoff
(\S\ref{sec:surprisal}), tested on three smaller models that share architecture and pipeline and differ
in the cutoff of their corpus (\S\ref{sec:series}), together with an account of what the probe measures (overlap with
the sources of the test text) and what it cannot resolve (\S\ref{sec:limits}).
\end{itemize}

\section{Related Work}

\paragraph{Vintage and chronologically consistent models.}
TypewriterLM~\citep{luo2026typewriter} is the closest work: a 7.2B model on 54B tokens of English from
1700--1913, with rule-based cleaning and an event-surprisal curve against a modern baseline. Talkie-1930~\citep{levine2026talkie}, Ranke-4B~\citep{goettlich2025history},
GPT-1900~\citep{hla2026machina} and TimeCapsuleLLM~\citep{grigorian2026timecapsule} are other models trained
from scratch; MonadGPT~\citep{langlais2023monadgpt} fine-tunes a modern model and so has no cutoff in our
sense, and \citet{underwood2025anachronism} argue that prompting and fine-tuning do not remove anachronism.
\citet{fittschen2026pretraining} pretrain small models on five periods of English, and
\citet{hosseini2021neural} released time-sliced models of nineteenth-century English before the current wave. In finance, model
families with yearly cutoffs are trained to control look-ahead
bias~\citep{he2026chronologically,drinkall2024time,yan2026datedgpt}. Outside English we know of three
community releases without papers: a 30M-parameter character model on a pre-1424 slice of
Kanripo~\citep{beltran2026ming}, whose author later found an eighteenth-century editorial layer in 19\% of
its documents (an independent instance of the problem of \S\ref{sec:corpus}; its repository held no
loadable weights when we tried to evaluate it), and two Japanese
models~\citep{zary2026edo,trtd2026timecapsuleja}.

\paragraph{Probing what a model knows of a period.}
That perplexity rises on text from after the training period, and that the rise should be read against a
control model, goes back to \citet{lazaridou2021mind} and \citet{dhingra2022time}. \citet{cheng2024dated}
date a model's effective cutoff from the perplexity of monthly Wikipedia revisions; TypewriterLM and Talkie
plot surprisal of dated events against modern baselines; \citet{vanadrichem2026year} estimate cutoffs of
base models without instructions. Our probe is of this family. What is specific to it is the test
material, long annals compiled after the cutoff whose entries are indexed by the year narrated, and the
controlled comparison of \S\ref{sec:series}. Text dating for Chinese has been studied as
classification~\citep{tian2021period,chang2021time}; we use an LLM on bibliographic information instead.

\paragraph{Classical Chinese models and benchmarks.}
\citet{chen2026internal} train a 318M model from scratch, mainly on Daizhige, and study whether it separates
real from invented events; they remove Latin letters and Arabic numerals but do not date the corpus, and no
weights are available. The UER GPT-2~\citep{zhao2019uer} is trained on the same collection.
SikuGPT~\citep{liu2023sikugpt} continues training GPT-2 checkpoints on the \emph{Siku quanshu} (completed
1782); encoders include SikuBERT and GuwenBERT~\citep{wang2022sikubert,ethanyt2020guwenbert}. Larger systems
are fine-tuned from modern models~\citep{yao2025wenyangpt,cao2024tonggu} or mix in modern
Chinese~\citep{li2025taiyan}. We report ACLUE~\citep{zhang2023aclue}, whose questions are posed in Modern
Chinese; WYWEB~\citep{zhou2023wyweb} is an alternative we did not run.

\section{Corpus: Making a Cutoff When Books Have No Dates}
\label{sec:corpus}

\subsection{The default outcome}
\label{sec:oldaudit}

Our first corpus used the obvious heuristic: assign each book the first dynasty name mentioned in its
opening lines, and keep books up to the target dynasty. The model trained on it, \old{}, was intended to
stop at 1840.

\emph{Corpus side.} In the 1.58B-token training file, late reign names occur thousands of times, as do
late-Qing and Republican figures and traces of modern editing (Table~\ref{tab:audit}). Matching eight late
markers per file flags 327 of 11{,}980 training files, of which 22 (4.8M characters) are dense enough to be
late works in their entirety.

\emph{Cause.} The heuristic read the surname \zh{梁} in the \emph{Collected Works of Liang Qichao}
(d.~1929) as the Liang dynasty (502--557), and \zh{元} in a personal name as the Yuan dynasty.
Heuristically dated books made up half of the training characters, and Qing text was about 1.5\% of the
corpus.

\emph{Model side.} Given the Qing reign names in order up to \zh{道光} (r.~1821--1850), the model's most
probable next character was \zh{咸}, the first character of the next reign name.

We report this failure because we think it is the default outcome: a model ``trained on Classical
Chinese'' has no cutoff unless one is built and then measured.

\begin{table}[t]
\small\centering
\begin{tabular}{@{}llrr@{}}
\toprule
Marker & & Old & New\\
\midrule
\zh{光緒} & reign, 1875--1908 & 2{,}633 & 0\\
\zh{宣統} & reign, 1909--1911 & 246 & 0\\
\zh{李鴻章} & statesman, d.~1901 & 707 & 0\\
\zh{王先謙} & scholar, d.~1917 & 288 & 0\\
\zh{王國維} & scholar, d.~1927 & 135 & 0\\
\zh{梁啓超} & writer, d.~1929 & 86 & 0\\
\zh{曾國藩} & statesman, d.~1872 & 77 & 0\\
\zh{出版社} & ``publishing house'' & 268 & 0\\
\zh{俞樾} & scholar, d.~1907 & 331 & 8$^\dagger$\\
\midrule
\zh{同治} & reign, 1862--1874$^\ast$ & 2{,}798 & 2{,}317\\
\zh{咸豐} & reign, 1851--1861$^\ast$ & 869 & 140\\
\bottomrule
\end{tabular}
\caption{Occurrences of late markers in the training file of \old{} (1.58B tokens) and \ours{} (1.24B).
These are the markers the cleaning rules use, so the New column shows that the rules were applied, not
that the corpus is clean (\S\ref{sec:postaudit}). Names of persons are candidate indicators of late material, not proof of it. $^\ast$Also ordinary phrases (e.g.\ \zh{諸味而同治}); we
inspected a sample of the remaining hits, not all. $^\dagger$Genuine residue: modern collation notes citing
Yu Yue.}
\label{tab:audit}
\end{table}

\subsection{Re-dating pipeline}
\label{sec:pipeline}

The candidates are 14{,}934 books (1.51B characters) from Daizhige and Kanripo. We define the cutoff by date
of composition: a book is admissible if its text was written before 1840, whenever it was first printed.

\textbf{(1) Canons are kept whole.} Texts filed under the Taish\=o canon, the Qianlong and Jiaxing canons
and the Zhengtong Daoist canon (5{,}291 books, 177M characters) are not re-dated at the book level; they
still go through step~4. This is the weakest step: the Taish\=o is a 1924--34 Japanese edition whose files
carry editorial matter (\S\ref{sec:postaudit}).

\textbf{(2) Book-level re-dating.} Each of the remaining 9{,}643 books, including those filed under
supplementary canons, is classified from its title, its category path in the collection, its length and
its first 160 characters as A, composed before 1840; B, composed later; C, a modern critical edition; or D,
undeterminable. Classification was done by an LLM (Claude Sonnet 5.5) in 33 batches; in six of them labels were
assigned by category rather than per title. The LLM outputs labels only; none of its output enters the
corpus.

\textbf{(3) Backstop rules.} A book labelled A is dropped if its text contains five or more late reign
names or three or more citations of modern scholars, or if step~4 would delete a thousand or more of its
paragraphs.

\textbf{(4) Paragraph-level cleaning.} In kept books, paragraphs containing reign names from Xianfeng
(1851) onward, publisher names, Gregorian years, modern collation formulae or modern scholars' names are
deleted (2.1M characters, 0.16\%, in 797 books).

\textbf{(5) Whole-book deduplication} by sampled clause hashes, larger books first: a book is dropped if
at least 70\% of its sampled clauses occur in a larger kept book (1{,}388 books, 70M characters).

\begin{figure}[t]
\centering
\includegraphics[width=\columnwidth]{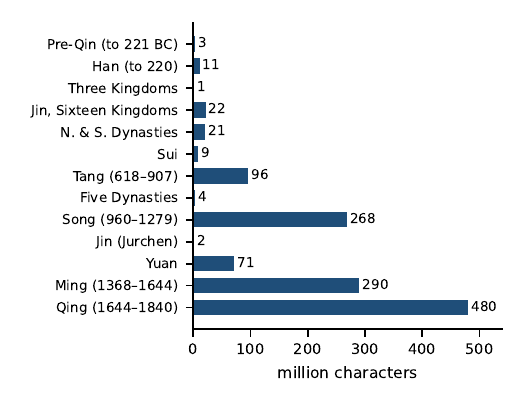}
\caption{Characters by period in the final corpus, by the collections' dynasty labels (not revised inside
the canons).}
\label{fig:corpus}
\end{figure}

Table~\ref{tab:pipeline} gives the outcome and Figure~\ref{fig:corpus} the distribution by period. For
training, books are resampled toward fixed shares per bibliographic division (with pre-Han texts repeated
up to 2.5 times and the canons downsampled by half), giving 1.24B training tokens and 6.8M validation
tokens, split by file.

\begin{table}[t]
\small\centering
\begin{tabular}{@{}lrr@{}}
\toprule
Decision & Books & M chars\\
\midrule
Canon, kept whole & 5{,}291 & 177\\
A: before 1840, kept & 8{,}359 & 1{,}173\\
\midrule
B: composed after 1840 & 164 & 27\\
C: modern critical edition & 245 & 42\\
D: undeterminable & 825 & 43\\
A, dropped by backstop rules & 50 & 51\\
\midrule
Removed as duplicates & 1{,}388 & 70\\
\textbf{Final corpus} & \textbf{12{,}262} & \textbf{1{,}278}\\
\bottomrule
\end{tabular}
\caption{Outcome of re-dating for the 14{,}934 candidate books.}
\label{tab:pipeline}
\end{table}

\subsection{Audits, and what they still find}
\label{sec:postaudit}

\paragraph{Same-marker audit.} In the new training file the markers of Table~\ref{tab:audit} drop to zero, or,
for strings that are also ordinary phrases, to occurrences that we inspected and found unrelated. Since
these are the strings the rules delete, this only confirms that the rules ran.

\paragraph{Manual sample.} We inspected 150 books labelled A, 90 of them deliberately drawn from the Qing
portion and Kanripo, where errors are most likely. One was a probable error and one doubtful; rule~3 was
added in response and catches both. This is a risk-enriched development sample used to revise the rules;
it does not estimate the residual error rate of the final corpus.

\paragraph{Held-out markers.} We then searched the final corpus for markers the pipeline does \emph{not}
use. This finds real residue of several kinds (details in Appendix~\ref{app:residue}):
\begin{itemize}
\setlength\itemsep{0pt}
\item \emph{A rule gap for 1841--1850.} The reign-name rules start at Xianfeng, so prefaces dated to the
last ten years of Daoguang pass: 14 such dates in 11 books.
\item \emph{Prefaces dated only by cyclical year or with variant characters}, among them colophons of 1914
and 1921 in two books.
\item \emph{Whole books that are late}: at least four labelled A appear to have been composed after 1840
(one mentions the Taiping war).
\item \emph{Modern apparatus not stripped}: one medical classic retains 2{,}741 notes by its 1924 editor;
38 books contain circled footnote markers.
\item \emph{Japanese editorial colophons}: 15 dated 1841--1928 in 14 files, most in the Taish\=o and
supplementary canons.
\end{itemize}
The four books total 0.64M characters (0.05\% of the corpus); we have no total for the other kinds. The
curves of \S\ref{sec:verify} are measured on a model trained with these exceptions, whose effect we have
not isolated. It means that \ours{} is a model whose corpus was \emph{selected and cleaned for} an 1840 cutoff, with known
exceptions, and not a model that has read nothing later. The released book list marks the affected files,
and the released scripts include the missing rules.

\section{Model}
\label{sec:model}

\ours{} is a decoder-only Transformer: 24 layers, width 1920, 20 heads with grouped-query attention,
RMSNorm, SwiGLU and rotary positions as in Llama~\citep{touvron2023llama}, 983M parameters, context
1{,}024. The vocabulary is character-level, 27{,}850 types collected from the corpus; 0.16\% of corpus
characters fall outside it. We train for 30{,}000 steps of 196{,}608 tokens (about 4.8 passes) with Muon
for matrix parameters and AdamW for the rest, on one H100 in 21 hours. The best validation loss is 2.22 nats
(3.21 bits) per character.

\paragraph{Transcription layer.} Three properties of the source collections are inherited by the model.
First, Daizhige is distributed in simplified characters and we convert it to traditional script
automatically; one-to-many mappings are resolved wrongly in a visible fraction of cases (the quotative
\zh{云} appears as \zh{雲} some 14{,}000 times after \zh{詩}, \zh{書} or \zh{傳}; \zh{里} ``mile'' as \zh{裏}). Second, all
text carries modern punctuation and paragraphing, the work of twentieth-century editors. Third, the converter writes \zh{爲}, \zh{衆}, \zh{啓} where most modern editions write \zh{為}, \zh{眾}, \zh{啟}, and the
corpus contains no quotation marks, title marks, question marks or exclamation marks. The cutoff therefore
applies to the wording of the texts, not to their orthography or punctuation, and text in other
conventions is penalised for reasons unrelated to its content (\S\ref{sec:surprisal}).

\paragraph{The earlier model.} \old{} has the same architecture and recipe but a different corpus: besides
the dating errors of \S\ref{sec:oldaudit}, a quarter of its training tokens were synthetic
worked-arithmetic text from an unrelated experiment. The two models differ in more than cleaning, and we
use \old{} as a documented failure case, not as an ablation. Both models are from single training runs.

\section{Probing the Cutoff}
\label{sec:verify}

A cleaned corpus is evidence about the data. We also want evidence from the model, obtained without relying on
it anything: a base LM cannot report its own cutoff, and prompting conflates knowledge with instruction
following.

\subsection{Relative surprisal on later-compiled annals}
\label{sec:surprisal}

\paragraph{Design.} We take annalistic histories \emph{compiled after the cutoff} that narrate, year by
year, events before and after it, and measure surprisal as a function of the year narrated. The texts
themselves are not in the corpus. Their \emph{sources} largely are: the imperial annals of \emph{Qingshi
gao} derive largely from the Veritable Records, directly or through the State History Office drafts and
the \emph{Donghua lu}, and \emph{Ming tongjian} draws on the \emph{Ming shi}. No file in the corpus
carries the title of either test work, of the \emph{Donghua lu} or of another history of the Qing. The
probe therefore measures how much of the record of a given year a model was trained on, which is the
quantity a cutoff controls, and its size depends on the pairing of corpus and test text.

\paragraph{Data.} (i) The annals of \emph{Qingshi gao} (compiled 1914--27), covering 1645--1911: 3{,}025
paragraphs. They are by several hands; the chapters for 1736--1874, which contain the cutoff, are
drafted by a single compiler, Wu Tingxie, and revised by others. (ii) \emph{Ming tongjian} (Xia Xie, printed 1873), covering
1368--1644, with the author's critical notes removed and up to 12 paragraphs sampled per year: 3{,}317
paragraphs. Paragraphs are truncated to 256 characters (732k characters scored). The two works differ in
genre, so comparisons across 1644 are confounded with the change of text; our main statistic is computed
within \emph{Qingshi gao}.

\paragraph{Measure.} For a model $M$ and a set of paragraphs $S$ let
$h_M(S)=\sum_{p\in S}\mathrm{bits}_M(p)\,/\sum_{p\in S}|p|$, where $|p|$ counts the characters of the
scored span, punctuation included. We report $r_M(S)=h_M(S)/h_R(S)$ for a modern reference model $R$,
since absolute surprisal moves together for all models (Table~\ref{tab:bpc}), and summarise by
$r_M(S_{\mathrm{post}})/r_M(S_{\mathrm{pre}})$ with $S_{\mathrm{post}}$ the events after 1840 and
$S_{\mathrm{pre}}$ those up to 1820. All quantities are pooled over characters, so the ratio of $r_A$ to
$r_B$ on the same set is $h_A/h_B$ exactly. The test text is first mapped to the conventions of our
transcription (\S\ref{sec:model}: eight variant characters; quotation and title marks removed; question
and exclamation marks replaced by full stops), for all models; without this mapping the models trained
on our corpus are penalised unevenly across the text, and we give both versions. Intervals are from a
bootstrap over blocks of narrated time, the same resample being used for numerator and denominator. Years
are the default block; neighbouring years are not independent, so for the main statistic we give
decade blocks as well. No interval includes variation across test works.

\begin{figure*}[t]
\centering
\includegraphics[width=0.76\textwidth]{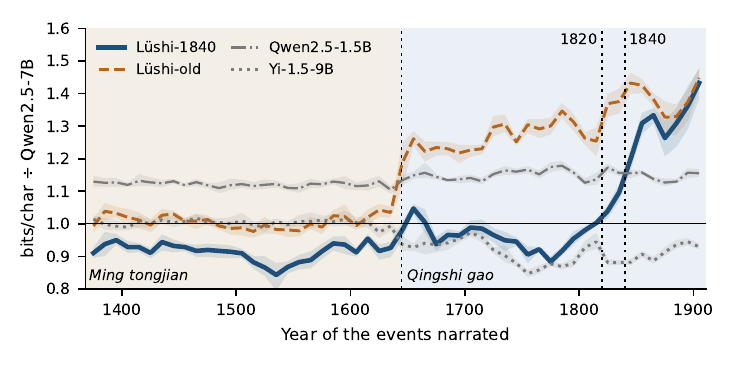}
\caption{Bits per character relative to Qwen2.5-7B by the year of the events narrated (10-year bins; bands
are 95\% intervals from a bootstrap over years). The test text changes at 1644, marked by the first dotted
line and the shading: \emph{Ming tongjian} (printed 1873) before, annals of \emph{Qingshi gao} (compiled
1914--27) after.}
\label{fig:surprise}
\end{figure*}

\paragraph{Results.} Within \emph{Qingshi gao}, the post/pre ratio of \ours{} against Qwen2.5-7B
\citep{yang2024qwen25} is 1.384, with a 95\% interval of 1.33--1.44 for decade blocks and 1.36--1.41 for
year blocks. Restricted to the chapters drafted by Wu Tingxie (1736--1820 against 1841--1874) it is
1.396; on the unmapped text it is 1.336 (1.310--1.362). Using any of four other modern models as the
reference gives 1.38--1.39 (Table~\ref{tab:refs}, top).

For four general-purpose models with no cutoff~\citep{yang2024qwen25,gemma2026gemma4,young2024yi} the same
statistic is 0.99--1.01 (Table~\ref{tab:refs}, bottom), and for a small GPT-2 trained almost entirely on
Classical Chinese Wikipedia~\citep{chang2024goldfish} it is 0.99. The rise is thus robust to the choice
among the modern references we tested; a bias shared by all of them is not excluded. Nor does this
contrast single out the cutoff: models of older text rise too, whether or not their corpus ends at 1840
(1.09 for \old{}; 1.09 for a GPT-2 trained on a collection that contains the test work; 1.09--1.20 for
three further models in \S\ref{sec:series} and \S\ref{sec:limits}). What is specific to the corpus has to
come from a controlled comparison.

By decade (Figure~\ref{fig:surprise}), the ratio of \ours{} moves between 0.84 and 1.05 for narrated years
from the 1360s to the 1810s. Within \emph{Qingshi gao}, from a low of 0.88 in the 1770s it rises almost monotonically: 0.95 in the
1790s, 1.04 in the 1820s, 1.10 in the 1830s, 1.20 in the 1840s and 1.31 in the 1850s, reaching 1.45 at the
end. This is a ramp, not a step, and it begins before the
nominal cutoff: the court chronicles that dominate our Qing material end with the Jiaqing reign (1820),
since those of Daoguang were compiled in 1856.

\begin{table}[t]
\small\centering
\setlength{\tabcolsep}{4pt}
\begin{tabular}{@{}lccc@{}}
\toprule
 & $\le$1820 & $>$1840 & post/pre [95\% CI]\\
\midrule
\multicolumn{4}{@{}l}{\emph{\ours{} divided by reference:}}\\
Qwen2.5-7B & 0.958 & 1.326 & 1.384 [1.359, 1.411]\\
Qwen2.5-1.5B & 0.831 & 1.158 & 1.394 [1.368, 1.420]\\
Gemma-4-E4B & 0.838 & 1.156 & 1.380 [1.353, 1.407]\\
Yi-1.5-6B & 1.030 & 1.428 & 1.386 [1.359, 1.412]\\
Yi-1.5-9B & 1.057 & 1.457 & 1.378 [1.350, 1.406]\\
\midrule
\multicolumn{4}{@{}l}{\emph{Model divided by Qwen2.5-7B:}}\\
Qwen2.5-1.5B & 1.153 & 1.145 & 0.993 [0.988, 0.998]\\
Gemma-4-E4B & 1.143 & 1.147 & 1.003 [0.998, 1.008]\\
Yi-1.5-6B & 0.930 & 0.929 & 0.999 [0.991, 1.007]\\
Yi-1.5-9B & 0.906 & 0.910 & 1.005 [0.994, 1.016]\\
Goldfish (lzh) & 1.600 & 1.582 & 0.989 [0.973, 1.006]\\
\old{} & 1.273 & 1.394 & 1.095 [1.079, 1.110]\\
GPT-2-DZG & 0.741 & 0.804 & 1.085 [1.057, 1.115]\\
\bottomrule
\end{tabular}
\caption{Relative bits per character within the annals of \emph{Qingshi gao}, for events of 1645--1820 and
1841--1911. Goldfish is scored on the unmapped text. GPT-2-DZG is the UER GPT-2 trained on Daizhige, which
includes \emph{Qingshi gao}.}
\label{tab:refs}
\end{table}

\subsection{A controlled cutoff series}
\label{sec:series}

To test the effect of extending the temporal coverage of the sources, we trained three 294M-parameter models (width 1024, 24 layers) with the tokenizer,
optimiser and schedule of \ours{}, each for 12{,}000 steps (2.36B tokens, about four hours on one H100), on
three corpora built by the pipeline of \S\ref{sec:corpus}. \emph{To-1644} drops every book labelled Qing
from the book list (773M training tokens); being defined by labels, it also loses Qing compilations of
earlier text. \emph{To-1840} is the corpus of \ours{} (1.24B). \emph{Extended} is the same training file with
34.3M characters appended once (2.7\%): the court records of the five later reigns (the Veritable Records
of Daoguang, Xianfeng, Tongzhi and Guangxu and the \emph{Xuantong zhengji}, 19.7M; the last two were
finished after 1911) and 113 books that re-dating had placed in 1841--1911 (14.6M). Neither test work is
in any of the three.

\begin{figure}[t]
\centering
\includegraphics[width=\columnwidth]{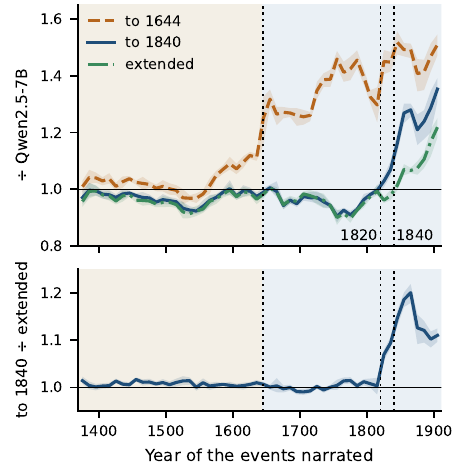}
\caption{The cutoff series: three 294M models differing in corpus. Top: relative to Qwen2.5-7B.
Bottom: the to-1840 model relative to the extended model, with no modern reference involved. Bins and
bands as in Figure~\ref{fig:surprise}.}
\label{fig:series}
\end{figure}

\paragraph{The rise moves with the corpus.} The to-1840 model has the curve of \ours{} at a third of the
size, with a somewhat smaller ratio: 0.96 before 1820 and 1.27 after 1840 within \emph{Qingshi gao}, a
post/pre ratio of 1.326 (1.303--1.348). For the extended model the ratio is 1.159 (1.139--1.182).
Dividing one by the other removes the modern reference (Figure~\ref{fig:series}, bottom): the two models
are within 0.99--1.02 of each other in every decade from the 1360s to the 1810s, and the ratio then steps
to 1.07 in the 1820s, 1.09 in the 1830s and 1.10--1.20 from the 1840s on (post/pre 1.144, 1.131--1.157).
A second training run of the to-1840 model gives 1.331 (1.308--1.353) against Qwen2.5-7B and stays within
0.99--1.03 of the first run in every decade; the extended model was trained once.

An earlier build, whose packing step had left the Daoguang (1821--1850) and Xianfeng (1851--1861)
records out of the training split, gives a further check: its model is 5--10\% above the extended model
in the 1820s to 1850s and within 0.97--1.01 of it in every other decade.

The position of the step is thus fixed by the design; the comparisons show that the probe responds to
the presence of sources in the decades they cover and not elsewhere. That the horizon of the to-1840 corpus itself lies near 1820 rests on the ramp of
\S\ref{sec:surprisal} and on a direct count: the share of six-character sequences of the test paragraphs
that occur in the to-1840 training file is 17\% for the annals up to 1820, 13\% for the 1820s, 10\% for the
1830s, 9\% for the 1840s and 2--4\% afterwards.

\paragraph{The to-1644 model} is at 0.97--1.12 by decade on \emph{Ming tongjian} and at 1.26--1.56 on
\emph{Qingshi gao} (post/pre 1.091, 1.074--1.108; overlap about 2\%). It has no rise at 1840 to speak of.
Its departure from the other two coincides with the change of test text and is already complete on the
annals for 1616--1643 (1.28), so it is not evidence of a horizon at 1644 (\S\ref{sec:limits}).

\paragraph{What remains unexplained.} The extended model is not flat: against Qwen2.5-7B it is at 1.01 in
the 1840s, 1.07 from the 1850s to the 1870s and 1.22 after 1900. Its coverage of the late period is thinner than of the
earlier one (overlap of 12\% after 1840 against 18\% before 1820, and 8\% for the
1860s), and the reference model may also know late-Qing events better than earlier ones; we cannot
separate the two. Under this setup, adding the later sources removes about half of the excess ratio of the to-1840 model
(1.326 to 1.159); we do not attribute the remainder. After 1840 the to-1644 model remains 17\% above the to-1840 model
though neither saw those years: familiarity with Qing official prose matters independently of events.

\subsection{What the probe does not show}
\label{sec:limits}

\paragraph{It does not date \old{} to 1644.} In Figure~\ref{fig:surprise} the curve of \old{} steps up at
1644, and its corpus does in effect end there (\S\ref{sec:oldaudit}). But the test text changes at the same
point. \emph{Qingshi gao} also has annals for the two pre-conquest rulers, covering 1616--1643 (221
paragraphs). On these \old{} stands at 1.27 relative to Qwen2.5-7B, as on the annals for
1645--1735 (1.25) and unlike its level on \emph{Ming tongjian} for the same years (1.04); \ours{} is at
1.01, 0.99 and 0.93. The step follows the work, or the Qing-compiled sources behind it, not the year
narrated, so this comparison cannot establish a 1644 horizon.

\paragraph{Similar curves do not identify a unique cutoff.} SikuGPT2-v1~\citep{lau2022sikugpt2v1} is an
early release of SikuGPT whose configuration shows it was initialised from a modern-Chinese GPT-2 and
further trained on the \emph{Siku quanshu}, an unpunctuated collection whose first set was completed in
1782. On the same paragraphs with everything but Han characters removed (for all models), its post/pre
ratio is 1.20 (1.18--1.22), against 1.39 (1.36--1.41) for \ours{}. Its ratio is level from the 1720s to
the 1810s and rises from the 1820s, the decade in which the rise of \ours{} accelerates. Because of its
initialisation this model has no known cutoff, so the comparison does not measure the resolution of the
probe. It shows that models with different corpora and training histories can produce curves of similar
shape, and that the shape alone does not date a model.

\paragraph{Memorisation looks like knowledge.} The UER GPT-2 trained on Daizhige, which contains
\emph{Qingshi gao}, has the lowest surprisal of all models on it (Table~\ref{tab:bpc}); its post/pre ratio is still 1.09.

\subsection{Courtesy names of historical persons}
\label{sec:zi}

Educated persons in pre-modern China had a courtesy name (\zh{字}), recorded in formulaic biographical
openings (``X, courtesy name Y''). From Wikidata we take 1{,}548 persons with a recorded courtesy name and
birth year (not filtered by nationality), sort them by birth year, form consecutive sets of four, and have
each model rank the four courtesy names after the prefix ``X\zh{，字}'' by log-probability.
Table~\ref{tab:zi} omits the 106 persons born before 1800 who died after it or have no recorded death year.

\begin{table}[t]
\small\centering
\setlength{\tabcolsep}{3pt}
\begin{tabular}{@{}lrcccc@{}}
\toprule
Persons & $n$ & Ours & Old & Q-1.5B & Q-7B\\
\midrule
died before 1800 & 1{,}172 & \textbf{.824} & .709 & .506 & .734\\
born 1800--1839 & 86 & .442 & .337 & .360 & \textbf{.581}\\
born 1840 or later & 184 & .348 & .364 & .435 & \textbf{.663}\\
\midrule
drop, first to last & & .476 & .345 & .071 & .071\\
\bottomrule
\end{tabular}
\caption{Top-1 accuracy on four-way courtesy-name items (Q: Qwen2.5). Bootstrap 95\% intervals over sets
of four, first and last rows: \ours{} [.799, .849], [.288, .408]; Qwen2.5-7B [.708, .758], [.582, .739].}
\label{tab:zi}
\end{table}

\ours{} falls by 0.476 (95\% interval 0.409--0.540) from the first group to the last; Qwen2.5-7B falls
by 0.071 ($-$0.012 to 0.155). For persons who died before 1800 the 983M-parameter \ours{} is more accurate
than the 7B general model (by 0.09, 0.06--0.12), and for those born after 1840 the order reverses
(by 0.32, 0.22--0.41). Intervals here resample the sets of four, since their members share distractors.
Two cautions apply. Uniform guessing gives 0.25, but a model without knowledge of the person can exceed
it through semantic and frequency cues, since courtesy names are related to given names. And the decline
is gradual: for the 40 persons born in 1760--1800 with known death years accuracy is already 0.43, since
biographies are written after the fact.

This probe does not transfer to the cutoff series of \S\ref{sec:series}. On persons born after 1840 the
to-1644, to-1840 and extended models score 0.33, 0.34 and 0.34, and the difference between the last two
is 0.00 ($-$0.06 to 0.06). The text added to the extended corpus is court records and assorted books,
which seldom give courtesy names, and at 294M parameters all three models are already at 0.36--0.42 for
persons born in the eighteenth century. The to-1644 model is below the to-1840 model even for persons
born before 1560 (by 0.05), since much of what is written about earlier persons is Qing. We therefore
read this probe as a diagnostic of what the model knows about persons, not as a test of the cutoff; about
0.35 is simply where all our from-scratch models end up.

\subsection{Reign-name continuation}
\label{sec:reign}

The probe that exposed the problem in \S\ref{sec:oldaudit} also documents the repair
(Table~\ref{tab:reign}). The first prompt is clean; the second and third themselves contain post-1840
elements (the temple name \zh{宣宗}, conferred in 1850, and the year Daoguang~30) and are shown for
completeness.

\begin{table}[t]
\small\centering
\begin{tabular}{@{}lrr@{}}
\toprule
Prompt (ending) & \old{} & \ours{}\\
\midrule
\ldots\zh{嘉慶、道光、} & 1 / $-$1.99 & 203 / $-$7.77\\
\ldots\zh{宣宗道光，文宗} & 77 / $-$6.64 & 337 / $-$10.49\\
\ldots\zh{道光三十年，} & 5 / $-$3.16 & 357 / $-$9.51\\
\bottomrule
\end{tabular}
\caption{Rank and log-probability of \zh{咸} (first character of the reign name Xianfeng, 1851--61) as the
next character. Qwen2.5-1.5B ranks it first under all three prompts.}
\label{tab:reign}
\end{table}

This is the only one of our probes that detects the late-Qing contamination of \old{}. Contamination
that is thin relative to a model's ignorance of a period is hard to see with knowledge probes, and none
of ours would detect the residue listed in \S\ref{sec:postaudit}.

\section{Capability}
\label{sec:capability}

On the two histories, for events up to 1820 and after the mapping of \S\ref{sec:surprisal}, \ours{} has
lower surprisal than Qwen2.5-1.5B (ratio 0.82), Gemma-4-E4B (0.81) and Qwen2.5-7B (0.93), and is close to
Yi-1.5-9B (0.96). On the unmapped text the ratio to Qwen2.5-7B is 1.015. The comparison favours \ours{}:
the sources of the test texts are in its corpus and the text is mapped to its conventions; the general
models are fed simplified script, and Qwen2.5-7B is an instruction-tuned checkpoint.
On ACLUE~\citep{zhang2023aclue}, which is posed in Modern Chinese, \ours{} reaches 34.1\% against 38.3\%
for Qwen2.5-1.5B (chance 25\%; Appendix~\ref{app:aclue}).

\section{Conclusion}

For pre-modern Chinese, a knowledge cutoff cannot be read off metadata; it has to be constructed book by
book and checked with instruments other than those used to build it, and ours still contains some later
material. Relative surprisal on later-compiled annals separates \ours{} from modern models, but
older-text models without an 1840 cutoff rise as well. In a controlled series, a model given the later
court records departs from the 1840 model in the decade where those records begin and nowhere before,
and adding them removes about half of the excess ratio. The probe measures overlap with the sources of
the test text: on it the boundary of coverage lies nearer 1820 than the nominal 1840, and the shape of a
curve alone does not date a model.

\section*{Limitations}

We release a base model only; there is no instruction-tuned version. Each model was trained once, except the to-1840 model of the cutoff series, which was trained twice; the three models of that series also differ in corpus size and so in the number of passes over it, the extended corpus was rebuilt twice after errors in its packing (the first build held the Daoguang records twice, lacked those of Xianfeng and included post-1911 books; the second, used above as a check, was also re-sampled throughout; the reported one appends the additions to the to-1840 file), and the point at which it departs from the 1840 corpus is fixed by what was added. The mapping of test text to the conventions of our transcription was introduced after a first round of scoring; both versions are reported. \old{}
and \ours{} differ in corpus size, period mix and the presence of synthetic text as well as in cleaning.
Book-level dating relied on an LLM reading bibliographic information, partly at the level of categories,
and was checked on a non-random sample of 150 books; the corpus is known to contain the residue described
in \S\ref{sec:postaudit}, and the released model was trained on it. We did not use existing authority files
(author dates in CBDB or the Kanripo catalogue), which would give a baseline for the LLM step. The four
canons were not dated at the book level. The 825 books we could not date are excluded, which biases the
corpus against poorly documented genres; the corpus is weighted toward official historiography, classical
commentary and the Buddhist and Daoist canons, and it includes vernacular fiction and translated
scripture as well as Literary Chinese proper. The text is a modern transcription with converted script and
modern punctuation. The surprisal probe rests on two works, only one of which spans the cutoff, and on
sources that are in the training corpus; its intervals reflect sampling of years, not of models or works.
The courtesy-name items come from Wikidata and inherit its coverage, including the fame bias of persons
born after 1840.

\section*{Ethical Considerations}

The training texts are pre-modern works in the public domain, taken from open transcriptions; the test
histories are also in the public domain. The model reproduces the outlook of its sources, including views
on gender, ethnicity and social hierarchy that are offensive today, and it is not a reliable source of knowledge about the modern
world; it is intended for research on historical language and on knowledge cutoffs. An LLM was used to
classify books by date (\S\ref{sec:pipeline}); AI assistants were also used for coding and for drafting
text, and the authors checked the results and take responsibility for the content.

\bibliography{refs}

\begin{thebibliography}{39}
\providecommand{\natexlab}[1]{#1}

\bibitem[{{Beltran12138}(2026)}]{beltran2026ming}
{Beltran12138}. 2026.
\newblock ming-vintage-scratch-30m: A from-scratch vintage {C}lassical
  {C}hinese language model.
\newblock Hugging Face model card,
  \url{https://huggingface.co/Beltran12138/ming-vintage-scratch-30m}.
\newblock Card revised 2026-08-24; accessed 2026-10-10.

\bibitem[{Cao et~al.(2024)Cao, Peng, Zhang, Shi, Liu, Ding, and
  Jin}]{cao2024tonggu}
Jiahuan Cao, Dezhi Peng, Peirong Zhang, Yongxin Shi, Yang Liu, Kai Ding, and
  Lianwen Jin. 2024.
\newblock \href {https://doi.org/10.18653/v1/2024.findings-emnlp.243}
  {{TongGu}: Mastering classical {C}hinese understanding with
  knowledge-grounded large language models}.
\newblock In \emph{Findings of the Association for Computational Linguistics:
  EMNLP 2024}, pages 4196--4210, Miami, Florida, USA. Association for
  Computational Linguistics.

\bibitem[{Chang et~al.(2021)Chang, Shiue, Yeh, and Demberg}]{chang2021time}
Ernie Chang, Yow-Ting Shiue, Hui-Syuan Yeh, and Vera Demberg. 2021.
\newblock Time-aware {A}ncient {C}hinese text translation and inference.
\newblock In \emph{Proceedings of the 2nd International Workshop on
  Computational Approaches to Historical Language Change}.
\newblock ArXiv:2107.03179.

\bibitem[{Chang et~al.(2026)Chang, Arnett, Tu, and Bergen}]{chang2024goldfish}
Tyler~A. Chang, Catherine Arnett, Zhuowen Tu, and Benjamin Bergen. 2026.
\newblock \href {https://aclanthology.org/2026.lrec-1.300/} {Goldfish:
  Monolingual language models for 350 languages}.
\newblock In \emph{Proceedings of the Fifteenth Language Resources and
  Evaluation Conference}, pages 3750--3781, Palma de Mallorca, Spain. ELRA
  Language Resource Association.

\bibitem[{Chen et~al.(2026)Chen, Lian, Wu, Huang, Sasaki, Kouno, and
  Choi}]{chen2026internal}
Jiuting Chen, Yuan Lian, Hao Wu, Tianqi Huang, Hiroshi Sasaki, Makoto Kouno,
  and Jongil Choi. 2026.
\newblock \href {https://arxiv.org/abs/2604.14180} {Internal knowledge without
  external expression: Probing the generalization boundary of a classical
  {C}hinese language model}.
\newblock \emph{Preprint}, arXiv:2604.14180.

\bibitem[{Cheng et~al.(2024)Cheng, Marone, Weller, Lawrie, Khashabi, and
  Van~Durme}]{cheng2024dated}
Jeffrey Cheng, Marc Marone, Orion Weller, Dawn Lawrie, Daniel Khashabi, and
  Benjamin Van~Durme. 2024.
\newblock \href {https://arxiv.org/abs/2403.12958} {Dated data: Tracing
  knowledge cutoffs in large language models}.
\newblock In \emph{Conference on Language Modeling (COLM)}.

\bibitem[{Dhingra et~al.(2022)Dhingra, Cole, Eisenschlos, Gillick, Eisenstein,
  and Cohen}]{dhingra2022time}
Bhuwan Dhingra, Jeremy~R. Cole, Julian~Martin Eisenschlos, Daniel Gillick,
  Jacob Eisenstein, and William~W. Cohen. 2022.
\newblock Time-aware language models as temporal knowledge bases.
\newblock \emph{Transactions of the Association for Computational Linguistics},
  10:257--273.

\bibitem[{Drinkall et~al.(2024)Drinkall, Rahimikia, Pierrehumbert, and
  Zohren}]{drinkall2024time}
Felix Drinkall, Eghbal Rahimikia, Janet Pierrehumbert, and Stefan Zohren. 2024.
\newblock \href {https://doi.org/10.18653/v1/2024.findings-naacl.208} {{Time
  Machine GPT}}.
\newblock In \emph{Findings of the Association for Computational Linguistics:
  NAACL 2024}, pages 3281--3292, Mexico City, Mexico. Association for
  Computational Linguistics.

\bibitem[{{Ethan-yt}(2020)}]{ethanyt2020guwenbert}
{Ethan-yt}. 2020.
\newblock \href {https://github.com/ethan-yt/guwenbert} {{GuwenBERT}: A
  pre-trained language model for classical {C}hinese (literary {C}hinese)}.
\newblock GitHub repository.

\bibitem[{Fittschen et~al.(2026)Fittschen, Li, Lippincott, Choshen, and
  Messner}]{fittschen2026pretraining}
Elisabeth Fittschen, Sabrina~Xin Li, Tom Lippincott, Leshem Choshen, and Craig
  Messner. 2026.
\newblock \href {https://doi.org/10.18653/v1/2026.findings-eacl.241}
  {Pretraining language models for diachronic linguistic change discovery}.
\newblock In \emph{Findings of the Association for Computational Linguistics:
  EACL 2026}, pages 4627--4642, Rabat, Morocco. Association for Computational
  Linguistics.

\bibitem[{{garychowcmu}(2018)}]{garychowcmu2018daizhige}
{garychowcmu}. 2018.
\newblock 殆知阁古代文献 ({Daizhige} v2.0).
\newblock GitHub repository, \url{https://github.com/garychowcmu/daizhigev20}.

\bibitem[{{Gemma Team}(2026)}]{gemma2026gemma4}
{Gemma Team}. 2026.
\newblock \href {https://arxiv.org/abs/2607.02770} {Gemma 4 technical report}.
\newblock \emph{Preprint}, arXiv:2607.02770.

\bibitem[{G{\"o}ttlich et~al.(2025)G{\"o}ttlich, Loibner, Jiang, and
  Voth}]{goettlich2025history}
Daniel G{\"o}ttlich, Dominik Loibner, Guohui Jiang, and Hans-Joachim Voth.
  2025.
\newblock \href {https://github.com/DGoettlich/history-llms} {History {LLM}s}.
\newblock Technical report, University of Zurich and Cologne University.

\bibitem[{Grigorian and Yaghoobian(2026)}]{grigorian2026timecapsule}
Hayk Grigorian and Hamed Yaghoobian. 2026.
\newblock \href {https://doi.org/10.1145/3803784.3807554} {{TimeCapsule}:
  Generative hallucination as a method for historical sensemaking}.
\newblock In \emph{Proceedings of the 2026 Conference on Creativity and
  Cognition}, pages 229--238. Association for Computing Machinery.

\bibitem[{He et~al.(2026)He, Lv, Manela, and Wu}]{he2026chronologically}
Songrun He, Linying Lv, Asaf Manela, and Jimmy Wu. 2026.
\newblock \href {https://doi.org/10.1016/j.jfineco.2026.104374}
  {Chronologically consistent large language models}.
\newblock \emph{Journal of Financial Economics}, 185:104374.

\bibitem[{Hla(2026)}]{hla2026machina}
Michael Hla. 2026.
\newblock {Machina Mirabilis}.
\newblock Blog post, \url{https://michaelhla.com/blog/machina-mirabilis.html}.

\bibitem[{Hosseini et~al.(2021)Hosseini, Beelen, Colavizza, and
  Coll~Ardanuy}]{hosseini2021neural}
Kasra Hosseini, Kaspar Beelen, Giovanni Colavizza, and Mariona Coll~Ardanuy.
  2021.
\newblock Neural language models for nineteenth-century {E}nglish.
\newblock \emph{Journal of Open Humanities Data}, 7.
\newblock ArXiv:2105.11321.

\bibitem[{{JeffreyLau}(2022)}]{lau2022sikugpt2v1}
{JeffreyLau}. 2022.
\newblock {SikuGPT2-v1}.
\newblock Hugging Face model card,
  \url{https://huggingface.co/JeffreyLau/SikuGPT2-v1}.
\newblock Accessed 2026-10-10.

\bibitem[{Langlais(2023)}]{langlais2023monadgpt}
Pierre-Carl Langlais. 2023.
\newblock {MonadGPT}.
\newblock Hugging Face model card,
  \url{https://huggingface.co/Pclanglais/MonadGPT}.

\bibitem[{Lazaridou et~al.(2021)Lazaridou, Kuncoro, Gribovskaya, Agrawal,
  Li{\v{s}}ka, Terzi, Gimenez, de~Masson~d'Autume, Ko{\v{c}}isk{\'y}, Ruder,
  Yogatama, Cao, Young, and Blunsom}]{lazaridou2021mind}
Angeliki Lazaridou, Adhiguna Kuncoro, Elena Gribovskaya, Devang Agrawal, Adam
  Li{\v{s}}ka, Tayfun Terzi, Mai Gimenez, Cyprien de~Masson~d'Autume,
  Tom{\'a}{\v{s}} Ko{\v{c}}isk{\'y}, Sebastian Ruder, Dani Yogatama, Kris Cao,
  Susannah Young, and Phil Blunsom. 2021.
\newblock Mind the gap: Assessing temporal generalization in neural language
  models.
\newblock In \emph{Advances in Neural Information Processing Systems},
  volume~34.

\bibitem[{Levine et~al.(2026)Levine, Duvenaud, and Radford}]{levine2026talkie}
Nick Levine, David Duvenaud, and Alec Radford. 2026.
\newblock Introducing talkie: a 13{B} vintage language model from 1930.
\newblock Blog post, \url{https://talkie-lm.com/introducing-talkie}.

\bibitem[{Li et~al.(2025)Li, Hu, and Wang}]{li2025taiyan}
Shen Li, Renfen Hu, and Lijun Wang. 2025.
\newblock \href {https://arxiv.org/abs/2505.11810} {Efficiently building a
  domain-specific large language model from scratch: A case study of a
  classical {C}hinese large language model}.
\newblock \emph{Preprint}, arXiv:2505.11810.

\bibitem[{Liu et~al.(2023)Liu, Wang, Zhao, Hu, Wu, Lin, Shen, Li, Liu, Zhang,
  and Zhao}]{liu2023sikugpt}
Chang Liu, Dongbo Wang, Zhixiao Zhao, Die Hu, Mengcheng Wu, Litao Lin, Si~Shen,
  Bin Li, Jiangfeng Liu, Hai Zhang, and Lianzheng Zhao. 2023.
\newblock \href {https://arxiv.org/abs/2304.07778} {{SikuGPT}: A generative
  pre-trained model for intelligent information processing of ancient texts
  from the perspective of digital humanities}.
\newblock \emph{Preprint}, arXiv:2304.07778.

\bibitem[{Luo et~al.(2026)Luo, Shinnick, Griesshaber, Wang, Yu, Shi, Torr, and
  Lu}]{luo2026typewriter}
Xiaoxi Luo, Zachary Shinnick, Niclas Griesshaber, Yixuan Wang, Junchi Yu, Freda
  Shi, Philip Torr, and Yao Lu. 2026.
\newblock \href {https://arxiv.org/abs/2606.02991} {A language model from 1913:
  Pretraining on historical text}.
\newblock \emph{Preprint}, arXiv:2606.02991.
\newblock Accepted to EMNLP 2026.

\bibitem[{Tian and K{\"u}bler(2021)}]{tian2021period}
Zuoyu Tian and Sandra K{\"u}bler. 2021.
\newblock \href {https://doi.org/10.18653/v1/2021.latechclfl-1.19} {Period
  classification in {C}hinese historical texts}.
\newblock In \emph{Proceedings of the 5th Joint SIGHUM Workshop on
  Computational Linguistics for Cultural Heritage, Social Sciences, Humanities
  and Literature}, pages 168--177, Punta Cana, Dominican Republic (online).
  Association for Computational Linguistics.

\bibitem[{Touvron et~al.(2023)Touvron, Lavril, Izacard, Martinet, Lachaux,
  Lacroix, Rozi{\`e}re, Goyal, Hambro, Azhar, Rodriguez, Joulin, Grave, and
  Lample}]{touvron2023llama}
Hugo Touvron, Thibaut Lavril, Gautier Izacard, Xavier Martinet, Marie-Anne
  Lachaux, Timoth{\'e}e Lacroix, Baptiste Rozi{\`e}re, Naman Goyal, Eric
  Hambro, Faisal Azhar, Aurelien Rodriguez, Armand Joulin, Edouard Grave, and
  Guillaume Lample. 2023.
\newblock \href {https://arxiv.org/abs/2302.13971} {{LLaMA}: Open and efficient
  foundation language models}.
\newblock \emph{Preprint}, arXiv:2302.13971.

\bibitem[{{trtd56}(2026)}]{trtd2026timecapsuleja}
{trtd56}. 2026.
\newblock {TimeCapsuleLLM-ja-v4}.
\newblock Hugging Face model card,
  \url{https://huggingface.co/trtd56/TimeCapsuleLLM-ja-v4}.
\newblock Accessed 2026-10-10.

\bibitem[{Underwood et~al.(2025)Underwood, Nelson, and
  Wilkens}]{underwood2025anachronism}
Ted Underwood, Laura~K. Nelson, and Matthew Wilkens. 2025.
\newblock \href {https://arxiv.org/abs/2505.00030} {Can language models
  represent the past without anachronism?}
\newblock \emph{Preprint}, arXiv:2505.00030.

\bibitem[{van Adrichem et~al.(2026)van Adrichem, Bhaskar, Yang, Potts, and
  Huang}]{vanadrichem2026year}
Suze van Adrichem, Aditi Bhaskar, Diyi Yang, Christopher Potts, and Jing Huang.
  2026.
\newblock \href {https://arxiv.org/abs/2608.15507} {Do language models
  consistently encode the current year?}
\newblock \emph{Preprint}, arXiv:2608.15507.

\bibitem[{Wittern(2014)}]{wittern2014kanripo}
Christian Wittern. 2014.
\newblock \href {https://dh-abstracts.library.virginia.edu/works/1972}
  {{K}anripo and {M}andoku: Tools for distributed repositories of premodern
  {C}hinese texts}.
\newblock In \emph{Digital Humanities 2014: Conference Abstracts}, Lausanne,
  Switzerland.

\bibitem[{Yan et~al.(2026)Yan, Tang, Gao, Jiang, and Lu}]{yan2026datedgpt}
Yutong Yan, Raphael Tang, Zhenyu Gao, Wenxi Jiang, and Yao Lu. 2026.
\newblock \href {https://arxiv.org/abs/2603.11838} {{DatedGPT}: Preventing
  lookahead bias in large language models with time-aware pretraining}.
\newblock \emph{Preprint}, arXiv:2603.11838.

\bibitem[{Yang et~al.(2024)Yang, Yang, Zhang, Hui, Zheng, Yu, Li, Liu, Huang,
  Wei, Lin, Yang, Tu, Zhang, Yang, Yang, Zhou, Lin, Dang, Lu, Bao, Yang, Yu,
  Li, Xue, Zhang, Zhu, Men, Lin, Li, Tang, Xia, Ren, Ren, Fan, Su, Zhang, Wan,
  Liu, Cui, Zhang, and Qiu}]{yang2024qwen25}
An~Yang, Baosong Yang, Beichen Zhang, Binyuan Hui, Bo~Zheng, Bowen Yu,
  Chengyuan Li, Dayiheng Liu, Fei Huang, Haoran Wei, Huan Lin, Jian Yang,
  Jianhong Tu, Jianwei Zhang, Jianxin Yang, Jiaxi Yang, Jingren Zhou, Junyang
  Lin, Kai Dang, and 23 others. 2024.
\newblock \href {https://arxiv.org/abs/2412.15115} {Qwen2.5 technical report}.
\newblock \emph{Preprint}, arXiv:2412.15115.

\bibitem[{Yao et~al.(2025)Yao, Wang, Chen, and Zhao}]{yao2025wenyangpt}
Xinyu Yao, Mengdi Wang, Bo~Chen, and Xiaobing Zhao. 2025.
\newblock \href {https://doi.org/10.24963/ijcai.2025/927} {{WenyanGPT}: A large
  language model for classical {C}hinese tasks}.
\newblock In \emph{Proceedings of the Thirty-Fourth International Joint
  Conference on Artificial Intelligence, {IJCAI-25}}, pages 8339--8347.
  International Joint Conferences on Artificial Intelligence Organization.
\newblock Main Track.

\bibitem[{Young et~al.(2024)Young, Chen, Li, Huang, Zhang, Zhang, Wang, Li,
  Zhu, Chen, Chang, Yu, Liu, Liu, Yue, Yang, Yang, Xie, Huang, Hu, Ren, Niu,
  Nie, Li, Xu, Liu, Wang, Cai, Gu, Liu, and Dai}]{young2024yi}
Alex Young, Bei Chen, Chao Li, Chengen Huang, Ge~Zhang, Guanwei Zhang, Guoyin
  Wang, Heng Li, Jiangcheng Zhu, Jianqun Chen, Jing Chang, Kaidong Yu, Peng
  Liu, Qiang Liu, Shawn Yue, Senbin Yang, Shiming Yang, Wen Xie, Wenhao Huang,
  and 12 others. 2024.
\newblock \href {https://arxiv.org/abs/2403.04652} {Yi: Open foundation models
  by 01.{AI}}.
\newblock \emph{Preprint}, arXiv:2403.04652.

\bibitem[{{zary0}(2026)}]{zary2026edo}
{zary0}. 2026.
\newblock {Edo-Cutoff LLM}: A language model with an 1868 knowledge cutoff.
\newblock Hugging Face model card,
  \url{https://huggingface.co/zary0/edo-cutoff-1b}.
\newblock Accessed 2026-10-10.

\bibitem[{Zhang and Li(2023)}]{zhang2023aclue}
Yixuan Zhang and Haonan Li. 2023.
\newblock \href {https://arxiv.org/abs/2310.09550} {Can large language model
  comprehend {A}ncient {C}hinese? {A} preliminary test on {ACLUE}}.
\newblock In \emph{Proceedings of the Ancient Language Processing Workshop},
  pages 80--87, Varna, Bulgaria. INCOMA Ltd., Shoumen, Bulgaria.

\bibitem[{Zhao et~al.(2019)Zhao, Chen, Zhang, Zhao, Liu, Lu, Chen, Deng, Ju,
  and Du}]{zhao2019uer}
Zhe Zhao, Hui Chen, Jinbin Zhang, Xin Zhao, Tao Liu, Wei Lu, Xi~Chen, Haotang
  Deng, Qi~Ju, and Xiaoyong Du. 2019.
\newblock \href {https://doi.org/10.18653/v1/D19-3041} {{UER}: An open-source
  toolkit for pre-training models}.
\newblock In \emph{Proceedings of the 2019 Conference on Empirical Methods in
  Natural Language Processing and the 9th International Joint Conference on
  Natural Language Processing (EMNLP-IJCNLP): System Demonstrations}, pages
  241--246, Hong Kong, China. Association for Computational Linguistics.

\bibitem[{Zhou et~al.(2023)Zhou, Chen, Wang, Zhong, and Zhang}]{zhou2023wyweb}
Bo~Zhou, Qianglong Chen, Tianyu Wang, Xiaomi Zhong, and Yin Zhang. 2023.
\newblock {WYWEB}: A {NLP} evaluation benchmark for classical {C}hinese.
\newblock In \emph{Findings of the Association for Computational Linguistics:
  ACL 2023}.
\newblock ArXiv:2305.14150.

\bibitem[{王东波 et~al.(2022)王东波, 刘畅, 朱子赫
  et~al.}]{wang2022sikubert}
王东波, 刘畅, 朱子赫, et~al. 2022.
\newblock {SikuBERT} 与
  {SikuRoBERTa}:面向数字人文的《四库全书》预训练模型构建及应用研究.
\newblock \emph{图书馆论坛}, 42(6):31--43.

\end{thebibliography}

\appendix

\section{Residual Late Material Found by the Held-Out Audit}
\label{app:residue}

\paragraph{Japanese colophons.} A scan of the whole corpus for a Japanese era name from Tenp\=o~11 (1840)
to Sh\=owa followed by a year gives 44 matches, of which 29 are homonymous Chinese era names (e.g.\ the
sixth-century era \zh{天保}) or misreadings. The other 15, in 14 files, are colophons or prefaces by
Japanese copyists and editors: \zh{孔雀經真言等梵本} (1841); \zh{佛果圜悟禪師碧巖錄} (1859, two files);
\zh{大通方廣懺悔滅罪莊嚴成佛經} (1862); \zh{華嚴策林} (1891); \zh{華嚴經入法界品十八問答} (1901);
\zh{三國遺事} (1902); \zh{中華傳心地禪門師資承襲圖} (1910); \zh{西方陀羅尼藏中金剛族阿蜜哩多軍吒利法}
(1912); \zh{遼東志} (1912, two passages); \zh{翰苑} (1922); \zh{妙法蓮華經玄贊} (1926); \zh{俱舍論疏}
(1927); \zh{梵語千字文} (1928). Most are in files of the Taish\=o canon; two are in the supplementary
canon, one in extra-canonical Buddhist files and two in the history division. Undated editorial remarks
are not found by this scan.

\paragraph{Daoguang 21--30.} Fourteen dates of the form ``Daoguang 2$x$/30'' (1841--1850) occur in 11
books, mostly in prefaces to medical and devotional works.

\paragraph{Later prefaces and colophons.} Examples found by searching for late figures and for cyclical
dates near them: a 1921 colophon by Ye Dehui in \zh{說文繫傳}; a 1914 colophon in \zh{逸老堂詩話}; a note
dated Xianfeng~7 written with a variant character in \zh{孔氏祖庭廣記}; prefaces postdating 1850 in
\zh{串雅內外編} and \zh{州縣事宜}.

\paragraph{Whole books.} \zh{南朝金粉錄} (mentions the Taiping war), \zh{俗話傾談} (printed 1870),
\zh{金鐘傳正明集} and \zh{舊晉書九家輯本} (compiler born 1804) were labelled A and may be later than
1840. The evidence in parentheses concerns printing or the compiler, not composition, so we list them as
uncertain.

\paragraph{Modern apparatus.} \zh{黃帝內經太素} retains 2{,}741 notes beginning \zh{平按} by its 1924
editor; circled note markers occur 4{,}359 times in 38 books (2{,}146 in one annotated
\emph{Zizhi tongjian}); a few files retain fragments of modern prefaces or of the web pages they were
copied from.

\section{Surprisal Items: Construction Details}
\label{app:surprisal}
Both works are taken from Chinese Wikisource. \emph{Ming tongjian}: chapters 5--94; year headings are
either ``reign name + year'' or, within a chapter, a bare year number, and are mapped to Gregorian years
by a table of Ming reign starts; the author's \zh{攷異} notes are removed; curly quotation marks are
normalised to corner brackets. \emph{Qingshi gao}: chapters 4--25 (and 1--3 for \S\ref{sec:limits}); year
headings may carry a reign prefix and a sexagenary year; the opening of the first chapter of each reign,
which recounts the end of the previous one, is skipped. Seven paragraphs at the head of chapter~17 that
escaped this rule are excluded. Twenty-one years lack a heading of their own in our source and are merged
into the preceding year; none lies between 1818 and 1883, and excluding these years and the ones they
were merged into leaves the main statistic at 1.385. All paragraphs of \emph{Qingshi gao} are used. General-purpose models (\texttt{Qwen/Qwen2.5-7B-Instruct}, \texttt{Qwen/Qwen2.5-1.5B}, \texttt{01-ai/Yi-1.5-6B},
\texttt{01-ai/Yi-1.5-9B}, \texttt{google/gemma-4-E4B}) and \texttt{uer/gpt2-chinese-ancient} are fed simplified
script, and our models, \texttt{goldfish-models/lzh\_hant\_full} and \texttt{JeffreyLau/SikuGPT2-v1} traditional script; all are
scored on the same 256-character spans after a beginning-of-sequence token. The mapping applied before
scoring is \zh{為}$\to$\zh{爲}, \zh{眾}$\to$\zh{衆}, \zh{啟}$\to$\zh{啓}, \zh{勛}$\to$\zh{勳}, \zh{裡}$\to$\zh{裏}, \zh{群}$\to$\zh{羣}, \zh{床}$\to$\zh{牀},
\zh{峰}$\to$\zh{峯}; corner brackets and title marks deleted; \zh{？} and \zh{！} replaced by \zh{。}. In the corpus \zh{為} is
41 times rarer than \zh{爲}, and in the annals it is 1.4\% of characters before 1820 and 0.7\% after 1840,
which is why the unmapped scores understate the post/pre ratio. Overlap in \S\ref{sec:series} is counted
on Han characters only, after the same mapping. As a check that the test works themselves are not in the
corpus under another title, we repeated the count with sequences of 20 characters: for \emph{Qingshi gao}
0.3\% of them occur in the to-1840 training file before 1820 and 0.01\% after 1840; for \emph{Ming
tongjian} 7.6\%, with 14 paragraphs almost wholly present. The ones we traced are memorials that
\emph{Ming tongjian} quotes and that also stand in earlier compilations in the corpus. The statistics in this paper are produced by scripts in
the release.

\begin{table}[h]
\small\centering
\setlength{\tabcolsep}{2.5pt}
\begin{tabular}{@{}lccccc@{}}
\toprule
 & \multicolumn{1}{c}{\emph{MTJ}} & \multicolumn{4}{c}{\emph{Qingshi gao} annals}\\
\cmidrule(lr){2-2}\cmidrule(l){3-6}
 & 1368-- & 1645-- & 1736-- & 1821-- & 1841--\\
Model & 1644 & 1735 & 1820 & 1840 & 1911\\
\midrule
\ours{} & 4.59 & 4.56 & 4.06 & 4.67 & 6.11\\
\old{} & 5.05 & 5.77 & 5.64 & 5.98 & 6.42\\
\midrule
Qwen2.5-1.5B & 5.62 & 5.28 & 5.06 & 5.04 & 5.27\\
Gemma-4-E4B & 5.74 & 5.28 & 4.98 & 4.99 & 5.28\\
Qwen2.5-7B & 5.01 & 4.62 & 4.36 & 4.32 & 4.60\\
Yi-1.5-6B & 5.17 & 4.39 & 3.97 & 3.96 & 4.28\\
Yi-1.5-9B & 5.02 & 4.33 & 3.82 & 3.80 & 4.19\\
\midrule
GPT-2-DZG$^\dagger$ & 4.67 & 3.65 & 3.03 & 2.83 & 3.70\\
\bottomrule
\end{tabular}
\caption{Bits per character by period of the events narrated (\emph{MTJ}: \emph{Ming tongjian}).
Paragraphs per column: 3{,}317 / 1{,}010 / 925 / 227 / 863. $^\dagger$UER GPT-2 trained on Daizhige, which
includes \emph{Qingshi gao}.}
\label{tab:bpc}
\end{table}

\section{Training and Release Details}
\label{app:repro}
\paragraph{Architecture and optimisation.} SwiGLU feed-forward width is $8/3$ of the model width rounded up
to a multiple of 64 (5{,}120 for \ours{}); grouped-query attention uses 5 key--value heads for 20 query
heads (4 for 16 in the 294M models). Matrix parameters inside the blocks are trained with Muon (learning
rate 0.02, Nesterov momentum 0.95, five Newton--Schulz steps), embeddings, output head and norms with AdamW
(learning rate $3\times10^{-3}$, $\beta=(0.9, 0.95)$, weight decay 0.1); 400 warm-up steps, then cosine
decay to 10\% of the peak; batches of $16\times1{,}024$ tokens with 12 accumulation steps; linear layers in
the blocks are trained in FP8. Training windows are sampled at random and no seed is fixed, so two runs
differ in initialisation and data order. Validation loss is computed every 500 steps and the best
checkpoint is kept. There was no hyperparameter search.

\paragraph{What will be released.} (i) The weights of \ours{} and of the series models. (ii) The list of
all 14{,}934 candidate books with source, label (A--D or canon), the rule that kept or dropped each, and
the label given by the LLM (\texttt{claude-sonnet-5-5}, run in October 2026), so that the classification
step need not be rerun. (iii) The manifest of the
exact corpus the checkpoint was trained on, with file hashes, kept separate from the repaired list in
which the residue of Appendix~\ref{app:residue} is marked. (iv) Test paragraphs with year labels,
per-paragraph scores for every model, and the scripts that produce each number. We do not redistribute
the transcriptions, which remain with their sources.

\section{ACLUE by Task}
\label{app:aclue}

On ACLUE~\citep{zhang2023aclue}, scored by option log-likelihood, \ours{} reaches 34.1\% micro-average
(binomial 95\% interval $\pm$1.3), above \old{} (31.5) and below Qwen2.5-0.5B (36.9) and 1.5B (38.3), with
chance at 25\% (Table~\ref{tab:aclue}). It is ahead of Qwen2.5-1.5B on eight of fifteen tasks and far
behind on sentence segmentation; on poetry sentiment and loan characters it is below chance. ACLUE was
scored as distributed, without the mapping of \S\ref{sec:surprisal}. ACLUE questions are written
in Modern Chinese, which \ours{} has not seen; we regard the benchmark as a weak measure for such a model
in either direction and make no capability claim from it.

\begin{table}[h]
\small\centering
\setlength{\tabcolsep}{3pt}
\begin{tabular}{@{}lrcccc@{}}
\toprule
Task & $n$ & Ours & Old & Q-0.5B & Q-1.5B\\
\midrule
Segmentation & 500 & 69.2 & 52.6 & 92.8 & \textbf{93.8}\\
Poetry quality & 406 & 52.0 & \textbf{61.6} & 44.1 & 50.5\\
Couplets & 500 & 38.6 & 32.2 & 38.0 & \textbf{42.4}\\
Medical texts & 211 & 33.2 & \textbf{33.6} & 24.2 & 29.4\\
Sinology basics & 136 & \textbf{32.4} & \textbf{32.4} & 30.1 & 31.6\\
Poetry apprec. & 103 & \textbf{31.1} & 25.2 & 16.5 & 15.5\\
Named entities & 500 & 29.6 & \textbf{30.4} & 25.2 & 25.2\\
Polysemy & 500 & 29.4 & 30.4 & \textbf{33.0} & 31.0\\
Poetry context & 500 & \textbf{29.2} & 25.8 & 26.8 & 26.2\\
Phonology & 100 & 28.0 & 30.0 & 31.0 & \textbf{36.0}\\
Language knowl. & 249 & 27.7 & \textbf{29.7} & 24.9 & 26.9\\
Reading compr. & 101 & \textbf{27.7} & 19.8 & 24.8 & 25.7\\
Literature knowl. & 160 & 27.5 & 26.2 & 33.8 & \textbf{39.4}\\
Loan characters & 500 & 20.0 & 19.2 & 26.4 & \textbf{27.4}\\
Poetry sentiment & 500 & 17.2 & 10.8 & \textbf{32.4} & 31.2\\
\midrule
Macro average & & 32.8 & 30.7 & 33.6 & \textbf{35.5}\\
Micro average & 4{,}966 & 34.1 & 31.5 & 36.9 & \textbf{38.3}\\
\bottomrule
\end{tabular}
\caption{ACLUE accuracy (\%), zero-shot by length-normalised option log-likelihood. Q: Qwen2.5. Task names
are our translations.}
\label{tab:aclue}
\end{table}

\end{document}